\documentclass[runningheads]{llncs}

\usepackage{style/main}

\usepackage{graphicx}
\usepackage{booktabs}

\usepackage[accsupp]{axessibility}  % Improves PDF readability for those with disabilities.

\usepackage[ruled,vlined]{algorithm2e}
\usepackage{enumitem}
\usepackage{hyperref}

\usepackage{orcidlink}

\newcommand{\del}[1]{}
\begin{document}

	\title{Analytic Convolutional Layer: \\ A Step To Analytic Neural Network} 
	\titlerunning{Analytic Convolutional Layer}
	\author{Jingmao Cui \orcidlink{0009-0006-5019-4933} \and 
		Donglai Tao \orcidlink{0009-0009-0602-754X} \and
		Linmi Tao \orcidlink{0000-0001-7731-708X} \and
		Ruiyang Liu \and Yu Cheng
		} 
		
	% Replace with an abbreviated list of authors.
	\authorrunning{J. Cui et al.}
	
	% Replace with your institution list.
	\institute{Department of Computer Science and Technology, Tsinghua University\\ Beijing 100084, China \\
		\email{linmi@tsinghua.edu.cn}\\
		\url{http://www.cs.tsinghua.edu.cn} }
	
	\maketitle
	
	\begin{abstract}
        The prevailing approach to embedding prior knowledge within convolutional layers typically includes the design of steerable kernels or their modulation using designated kernel banks. In this study, we introduce the Analytic Convolutional Layer (ACL), an innovative model-driven convolutional layer, which is a mosaic of analytical convolution kernels (ACKs) and traditional convolution kernels. ACKs are characterized by mathematical functions governed by analytic kernel parameters (AKPs) learned in training process. Learnable AKPs permit the adaptive update of incorporated knowledge to align with the features representation of data. Our extensive experiments demonstrate that the ACLs not only have a remarkable capacity for feature representation with a reduced number of parameters but also attain increased reliability through the analytical formulation of ACKs. Furthermore, ACLs offer a means for neural network interpretation, thereby paving the way for the intrinsic interpretability of neural network. The source code will be published in company with the paper.
		
	\end{abstract}

	\section{Introduction}
	\label{sec:intro}
	
	\begin{figure}[tb]
		\centering
		\includegraphics[width=1\linewidth]{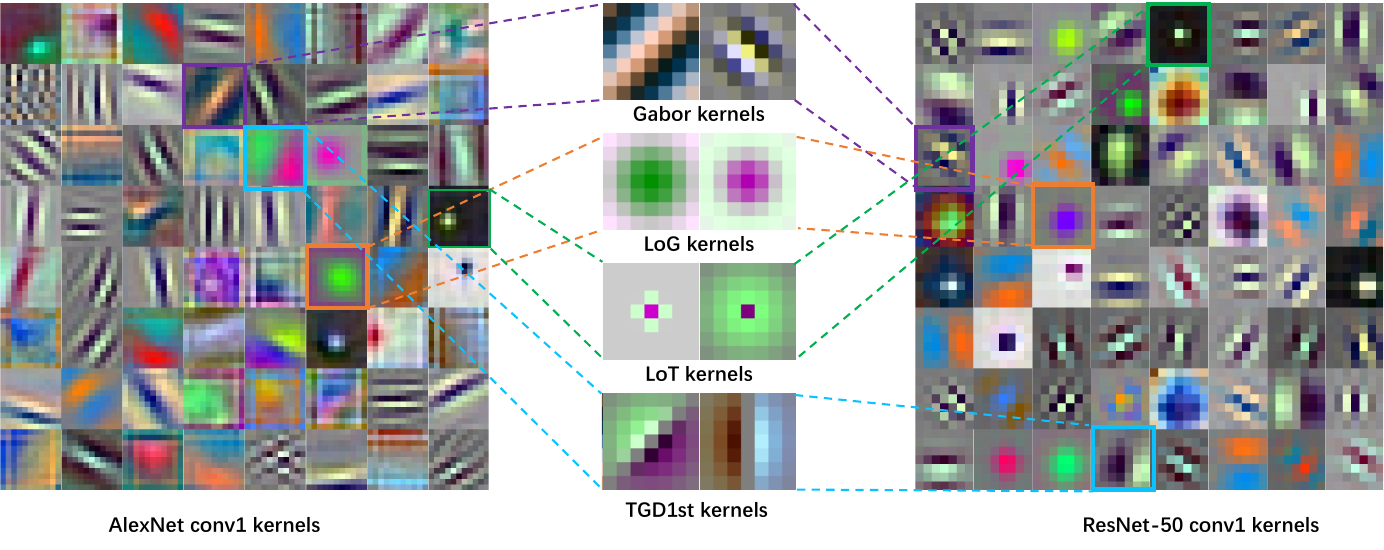}
	\caption{Convolution kernels pretrained on ImageNet exhibit discernible patterns, which we demonstrate can be analytically modeled using mathematically-defined kernels. The figure illustrates the first layer kernels from pretrained AlexNet \cite{krizhevsky2012imagenet} and ResNet-50 \cite{He2015DeepRL} on the left and right, respectively, with the corresponding analytical kernels in the center. It is important to note that these vibrant kernels represent combinations of three distinct kernels within the red, green, and blue channels, respectively.}  
		
		\label{fig:sim}
	\end{figure}
	
	With the dawn of the big model era, models like LLaMA \cite{touvron2023llama} boast 7 to 70 billion parameters, and GPT-4 \cite{openai_gpt-4_2023} reportedly houses a staggering 1.76 trillion parameters. The prowess of these mammoth models is undeniable; however, the problem of reliability and redundancy looms large in the research community. The crux of these challenges lies in the inherently data-driven nature of contemporary neural networks. In essence, these data-driven models suffer from a lack of intrinsic mechanisms to gracefully handle the training data, resulting in an enormous parameters.
	
	Enhancing the feature representation of neural networks often involves embedding prior knowledge into convolution kernels. Deformable kernels~\cite{dai2017deformable} enable the sampling grid to undergo free-form deformation, albeit with the introduction of additional parameters for learning offsets. Zhou \etal\cite{zhou2017oriented} developed Active Rotating kernels that rotate during the convolution process, thereby explicitly encoding the location and orientation in the resultant feature maps. Similarly, Luan \etal\cite{luan2018gabor} employed Gabor kernel banks to adjust convolution kernels across various orientations and scales. Despite these advancements, none of these approaches have succeeded in substantially reducing model parameters, as all convolution kernels still require updating. Furthermore, these methods retain static prior knowledge during the training process, without allowing for any adaptive changes. 
	
	Researchers have observed distinct patterns of convolution kernels within convolutional layers (\cref{fig:sim}) pretrained on large natural image datasets \eg ImageNet \cite{deng2009imagenet}, representing crucial \textit{a priori} knowledge that has spurred further investigation. Shang \etal~\cite{shang_understanding_2016} noted that kernels in the initial layers often form pairs, which are postulated to be redundant kernels for extracting both positive and negative phase information from input signals. Luan \etal\cite{luan2018gabor} identified that certain convolution kernels bear a resemblance to Gabor kernels and have thus utilized Gabor kernel banks to modulate convolution kernels accordingly. We observed that the majority of emergent patterns in pretrained lower-layer convolution kernels correspond to the characteristic patterns of specific kernel functions: these include Gabor\cite{fogel1989gabor}  kernels, Laplacian of Gaussian\cite{Sotak1989TheLK} (LoG) kernels, and TGD kernels as presented by Tao \etal\cite{tao2023tao}, encompassing both first-order (TGD1st) and second-order (TGD2nd) 2D TGD kernels, as well as Laplacian of TGD (LoT) kernels. Various types of analytic kernels serve disparate roles; for instance, Gabor kernels commonly characterize textures and oriented patterns, while LoG and LoT kernels are adept at detecting isotropic properties. Conversely, TGD1st kernels typically depict features with both positive and negative segments.
	
	Drawing on these observations, we have modeled convolution kernels with these kernel functions (hereafter referred to as analytic convolution kernels, ACKs), as depicted in \cref{fig:sim}. Convolutional layers that employ ACKs are designated as Analytic Convolutional Layers (ACLs). The weight of an ACL is a mosaic of two distinct components: ACKs and the traditional convolution kernels (labeled as plain kernels). The parameters governing ACKs are referred to as Analytic Kernel Parameters (AKPs), which are updated to more accurately reflect the features of the data space throughout the training phase. Recognizing that certain convolution kernels cannot be adequately represented by known kernel functions we advocate retaining some fraction of plain kernels in the ACLs, which undergo updates during the training process analogously to kernels in traditional convolutional layers. 
	
	AKPs are the cornerstone of ACLs, as they infuse diverse facets of \textit{a priori} knowledge into the neural network via distinct kernel functions, underscoring the model-driven nature of ACLs. ACLs preserve a robust capacity for representing feature space by iteratively refining AKPs during the backpropagation process, thereby dynamically assimilating characteristic information of the feature space, such as orientations, scales, phases, and edges. Such a strategy promises to yield not only more efficient models, possessing a condensed parameter profile but also maintaining their potent feature representation capabilities.
	
	The contributions of this paper are summarized as follows:
	
	\begin{itemize} [topsep=0pt] 
		\item[$\bullet$] This paper presents the first attempt to analytically model convolution kernels (ACK), and to construct the model-driven Analytic Convolutional Layer (ACL). 
		\item[$\bullet$] We provide derivations for the updates of AKPs within the learning cycles. It incorporates prior knowledge conveyed by AKPs into the kernels and updates during learning to enhance representation of features. 
		\item[$\bullet$] The ACL represents a foundational step toward building analytic neural networks. These layers are highly compatible with various network architectures, which offer a substantial possibility to the interpretability of neural network. 
	\end{itemize}
	
	\section{Related Work}
	\label{sec:relate}
	
	\subsection{Receptive field models of visual system}
	Cells in the mammalian primary visual cortex exhibit a diversity of receptive field shapes\cite{hubel_receptive_1962,de_valois_spatial_1982, bonin2005suppressive}, which served as the foundational inspiration for selecting kernel functions for analytic convolution kernels (ACKs). Neurons, with their uniform receptive fields unresponsive to the orientation of light stimuli, led to the adoption of mean value kernels. The center-surround receptive fields of retinal ganglion neurons and lateral geniculate nucleus cells, displaying isotropic features, steered the choice towards LoG kernels and LoT kernels. The strip- or edge-like receptive fields of simple cells in primary visual cortex, which show sensitivity to orientation, informed the decision to utilize Gabor kernels. Conversely, complex cells feature bar-shaped receptive fields and an orientation sensitivity, indicative of the use of TGD1st kernels. The receptive fields of super-complex cells, encompassing higher-order patterns, remain an area of exploration for future kernel functions. 
	
	\subsection{Tao General Difference}
	Tao \etal\cite{tao2023tao} initially presented a rigorous mathematical expression for the differentiation of discrete sequences through their Tao General Difference (TGD) theory. TGD operators have demonstrated exceptional signal processing capabilities, including robust noise resistance, positioning them as an optimal choice for edge detection in natural images. In this study, we implemented three types of TGD kernels; namely, first-order TGD (TGD1st) kernels, second-order TGD (TGD2nd) kernels, and Laplacian of TGD (LoT) kernels. TGD1st kernels, which consist of positive and negative components, offer a compelling alternative to Gabor kernels for replicating the receptive fields of simple and complex cells. Additionally, TGD2nd kernels have applications in edge detection, and LoT kernels excel at identifying isotropic features.
	
	\subsection{Convolutional layer design}
	Researchers have persistently endeavored to infuse prior knowledge into neural networks by creating specialized convolutional layers, aiming for enhanced accuracy, reduced parameters, and deeper insights into neural network operations. Yao \etal\cite{yao2016gabor} directly employed outputs from Gabor kernels as inputs to CNNs, while Sarwar \etal\cite{sarwar2017gabor} incorporated Gabor kernels into the first or second convolutional layer. However, these Gabor kernels remain static and do not update during the backpropagation (BP) process, limiting their adaptability to specific feature spaces. Luan \etal\cite{luan2018gabor} utilized predefined Gabor kernel banks to modulate convolution kernels across any layer, yet their approach does not effectively diminish the number of training parameters. In contrast, our ACLs select ACKs from a diverse pool, including Gabor, LoG, TGD1st, TGD2nd, LoT kernels, and others, offering greater flexibility and functionality. More crucially, AKPs of the ACKs are dynamically updated via the BP algorithm, ensuring that the prior knowledge encapsulated by the AKPs continuously evolves to align with the feature space of the training dataset.

	\section{Analytic Convolutional Layer}
	\label{sec:acn}
	
	\begin{figure}[tb]
		\centering
		\includegraphics[width=1.\linewidth]{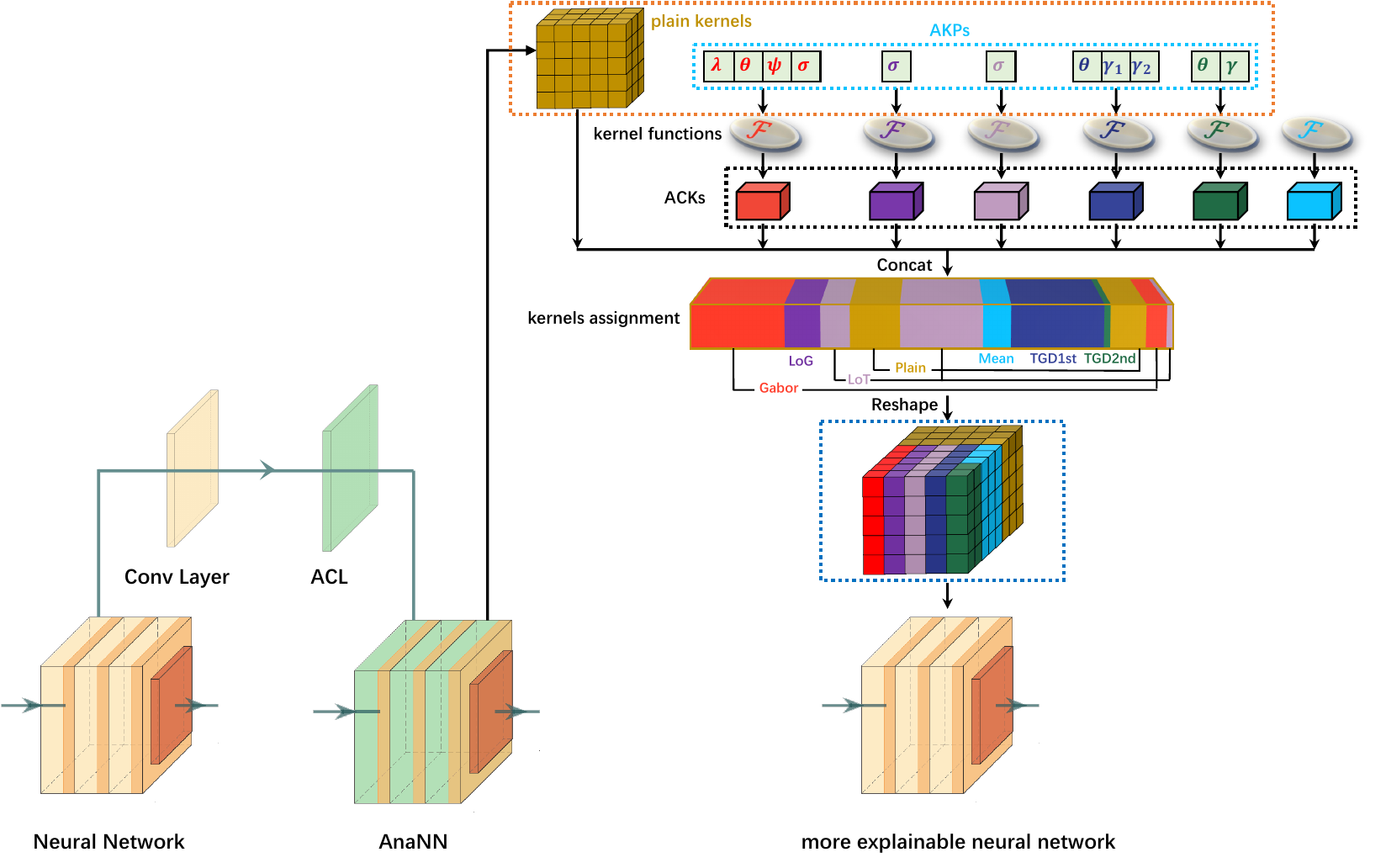}
		\caption{The diagram illustrates the workflow of the ACL  and delineates the process of transforming a standard neural network into an AnaNN (Analytic Neural Network) to enhance its explainability. The parameters enclosed within the orange dashed box, which include plain kernels and AKPs, are the only ones that are learned and updated. The blue dashed box represents the typical volume of parameters that must be learned and updated in traditional convolutional layers.}
		\label{fig:AnaNN}
	\end{figure}
	
	\subsection{Model-driven convolutional layer}
	
	In contrast to the data-driven approach of traditional convolutional layers, ACL adopts a model-driven strategy that posits convolution kernels should perform distinct functions, akin to the specialized cells within the human visual system. With ACL, practitioners have the discretion to specify the number of different types of ACKs to employ, the sequence of their arrangement, and the relative proportions between ACKs and plain convolution kernels. This substantial flexibility grants the ACL the capability to encapsulate complex, larger feature spaces effectively.
	
	\subsection{AConv}
	
	The convolution operation within an ACL is referred to as AConv. An ACL with $ C_i $ input channels and $ C_o $ output channels comprises a total of $ C_o \times C_i $ convolution kernels. The size of each kernel is denoted as $ h \times w $, and the weight of the $ k $-th kernel\del{, which can be an ACK or plain convolution kernel,} is represented by the matrix $ \mathbf{W}^k_{h \times w} $. When the $ k $-th kernel is associated with the $ p $-th input channel and the $ q $-th output channel, we can equate the notation $ \mathbf{W}^{pq} $ with $ \mathbf{W}^k $, and this equivalence extends to other related notations. Whereas in traditional convolutional layers $ \mathbf{W}^k $ is commonly trainable on an element-wise basis, our ACL employs a distinct \emph{kernel function}, $ f^k $, to govern $ \mathbf{W}^k $ for each respective kernel. 
	
	\begin{definition}
		A kernel function \( f^k \) with \( n \) parameters is represented in the abstract form:
		\begin{equation}
			f^k(x, y; \phi^k_1, \phi^k_2, \ldots, \phi^k_n)
		\end{equation}
		This function is defined over the product space \( \left(-\frac{h}{2}, \frac{h}{2}\right) \times \left(-\frac{w}{2}, \frac{w}{2}\right) \).
		
		The matrix \( \mathbf{W}^k \) consists of the sampled values of \( f^k \) on a grid that corresponds to a \( h \times w \) shape, delineated as follows:
		
		\begin{align}
				&& \mathbf{W}^k_{ij} &= f^k\left(i', j'; \phi^k_1, \ldots, \phi^k_n\right), \\
		where: && i' &= i - \frac{h+1}{2}, \text{~} j' = j - \frac{w + 1}{2} \\
				\text{for } && i &= 1, 2, \ldots, h \text{ and } j = 1, 2, \ldots, w. \notag
		\end{align}

	\end{definition}
	
	Given a predefined kernel function, the corresponding weight matrix can be refined by altering the values of $(\phi_1, \ldots, \phi_n)$, which are henceforth designated as analytic kernel parameters (AKPs). A plain kernel (traditional convolution kernel) may be viewed as a special example of ACKs, with its kernel function formulated as: 
	\begin{equation}
		f^k(x,y; \phi^k_1, \phi^k_2, \dots, \phi^k_{h \times w}) = \left\{\begin{aligned}
			\phi^k_{(i-1)w + j}, && x = i', y = j' \\
			0, && otherwise.
		\end{aligned}\right.
	\end{equation}
	Therefore, ACKs can contain up to $h \times w$ parameters,  equating to the parameter count found in a traditional convolution kernel. The structure of the kernel function dictates its fundamental behavior, while AKPs imbue it with adaptability. Consequently, the kernel function for each kernel is predefined during the initial layer design phase, with AKPs retained as trainable variables.
	
	Besides the previously described kernels, an ACL may include a bias vector $\mathbf{b}_{C_o \times 1}$ and other potential hyperparameters, such as stride, padding method, and dilation. These hyperparameters adhere to the same constraints and serve identical functions as those in traditional convolutional layers.
	
	\textbf{Forward Pass} The forward pass of an ACL closely mirrors that of a traditional one. Given an input feature $\mathbf{F}$ of dimensions $(B, C_i, H, W)$, we initially deduce the kernel weight matrices $\mathbf{W}^k$ from the analytic kernel parameters (AKPs) $\phi^k$. Subsequently, the resultant output feature $\mathbf{F}'$, which possesses dimensions $(B, C_o, H', W') $, is calculated by
	
	\begin{equation}
		\mathbf{F}'_{mq} = \mathbf{b}_q + \sum_{p=1}^{C_i} \mathbf{W}^{pq} \star \mathbf{F}_{mp}
	\end{equation}
	where $\star$  denotes the valid 2D cross-correlation operation.

	\textbf{Backpropagation} In an ACL, the weight matrix $\mathbf{W}^k$ is exclusively determined by the kernel function $f^k$ and its associated AKPs $\phi^k_1, \ldots, \phi^k_n$. Consequently, during the backpropagation, the gradients of the AKPs are aggregated. Given a loss function $\mathcal{L}$, the gradient $\delta^k_l$ for an AKP $\phi^k_l$ is expressed as
	\begin{align}
	    \delta^k_l = \frac{\partial \mathcal{L}}{\partial \phi^k_l} &= \sum_{i, j} \left( \frac{\partial \mathcal{L}}{\partial \mathbf{W}^k_{ij}} \frac{\partial \mathbf{W}^k_{ij}}{\partial \phi^k_{l}} \right) \\
	     &= \sum_{i, j} \left( \frac{\partial \mathcal{L}}{\partial \mathbf{W}^k_{ij}} \left.\frac{\partial f^k(x, y; \phi^k_1, \ldots, \phi^k_n)}{\partial \phi^k_l}\right|_{x = i', y = j'} \right)
	\end{align}
	
	Subsequently, the update to $\phi^k_l$ is performed using the rule
	
	\begin{equation} \phi^k_l \leftarrow \phi^k_l - \eta \delta^k_l \end{equation}
	
	where $\eta$ represents the learning rate.
	
	\subsection{Kernels arrangement}
	
	Unlike the equal status typically afforded to individual convolution kernels within a traditional convolutional layer, ACKs in an ACL assume distinct roles, rendering their status unequal. The order of these kernels is crucial, as their relative positioning contributes significantly to the representational capacity of the ACL.
	
	We use $\mathcal{N}{i}$ to denote the $i$-th component of  the weight matrix of an ACL, which may represent any type of ACKs or plain convolution kernels. Let $n_{i}$ indicate the quantity of each component. This arrangement yields a pattern that uniquely identifies an ACL:
	\begin{equation}
		\label{eq:unique}
		\mathbf{(C_{i} \times C_{o})\mathcal{N}_{1(n_{1})}\mathcal{N}_{2(n_{2})} \dots \mathcal{N}_{m(n_{m})}}
	\end{equation} 
	where all $n_{i}$ should sum to $C_{i} \times C_{o}$ . \cref{eq:unique} tells this ACL have $C_{i}$ in-channels and $C_{o}$ out-channels, the kernels are composed of  $n_{1}$ kernels of type $\mathcal{N}_{1}$, $n_{2}$ kernels of type $\mathcal{N}_{2}$,\dots,$n_{m}$ kernels of type $\mathcal{N}_{m}$ in order. All $\mathcal{N}_i$ can be the same or totally different. The ratio form of \cref{eq:unique} is more handy in deeper networks:
	\begin{equation}
		\label{eq:uniqueRatio}
		\mathbf{(C_{i} \times C_{o})\mathcal{N}_{1(r_{1})}\mathcal{N}_{2(r_{2})} \dots \mathcal{N}_{m(r_{m})}}
	\end{equation}  
	where $r_{i}$ is the ratio of $n_{i}$ to $C_{i} \times C_{o} $. Without ambiguity, we can omit $C_{i}$ and $C_{o}$.
	
	\subsection{Analytic kernel functions}
	
	Our ACL framework requires only one final element to achieve completion: the selection of appropriate kernel functions to construct analytic models. The fidelity of these models is paramount, as it determines the ACL's capacity to accurately represent the feature space. Drawing on the insights gleaned from the receptive fields of visual system cells and observations presented in Figure \ref{fig:sim}, we introduce a suite of user-friendly analytic kernel functions.
	
	\textbf{Mean kernel function}  Mean kernels (short for mean value kernel) can be get by:
	\begin{equation}
		\mathit{Mean} = \mathbf{M}/\sqrt{h \times w}
	\end{equation}
	Where $\mathbf{M}$ denotes an all-one matrix of dimensions $h \times w$. Mean kernels, possessing no AKPs, remain static; they neither learn nor update during the training process when employed as ACKs. Consequently, this reduces the number of parameters by $h \times w$, as well as the computational load. Given their simplicity and inability to encapsulate specific information within the feature space, mean kernels are aptly utilized in ablation studies to assess the functionality of other ACKs.
	
	\textbf{Gabor kernel function}
	We adopt the real part of Gabor function\cite{fogel1989gabor} to build a Gabor kernel:
	\begin{align}
		&& \mathit{Gabor}(x, y ; \lambda, \theta, \psi, \sigma, \gamma) &= \exp \left(-\frac{x^{\prime 2}+\gamma^2 y^{\prime 2}}{2 \sigma^2}\right) \cos \left(2 \pi \frac{x^{\prime}}{\lambda}+\psi\right)\\
where:	&& x^{\prime} &= x \cos\theta	+ y \sin\theta\ \\
		&& y^{\prime} &= -x \sin\theta	+ y \cos\theta
	\end{align}

	Where $\theta$ denotes the orientation, $\lambda$ signifies the wavelength of the sinusoidal component, $\psi$ represents the phase offset, $\sigma$ describes the Gaussian envelope's standard deviation, and $\gamma$ indicates the spatial aspect ratio, determining the ellipticity of the Gabor function's support. In our subsequent experiments, we fix $\gamma$ to 1, thereby reducing the Gabor kernel to four AKPs.
	
	\textbf{LoG kernel function}
	Laplacian of Gaussian with standard deviation $\sigma$ takes the form:
	\begin{equation}
		\mathit{LoG}(x, y;\sigma)=-\frac{1}{\pi \sigma^4}\left[1-\frac{x^2+y^2}{2 \sigma^2}\right] e^{-\frac{x^2+y^2}{2 \sigma^2}}
	\end{equation}
	The LoG kernel is isotropy, it has only one AKP $\sigma$ controlling the deviation.
	
	\textbf{TGD1st kernel function}
	TGD kernels may be implemented through various approaches, contingent upon the selection of kernel functions \cite{tao2023tao}. For simplicity, we opt for two exponential functions corresponding to the two dimensions. Consequently, our TGD1st kernel function adopts the following form:
	\begin{align}
		&&\mathit{TGD1st}(x,y;\theta, \gamma_1, \gamma_2)  &= e^{- \frac{x^{\prime 2}}{\gamma_1^2} - \frac{y^{\prime 2}} {\gamma_2^2} } \times sign(\cos\theta) \times \mathbf{I}_{|x^{\prime}|>\delta} \\
where:  && x^{\prime} &= x \cos\theta	+ y \sin\theta \\
		&& y^{\prime} &= -x \sin\theta	+ y \cos\theta
	\end{align}

	In this context, the indicator function $\mathbf{I}_{|x^{\prime}|>\delta}$ is defined as 1 when $|x^{\prime}| > \delta$, and 0 otherwise, effectively distinguishing the positive and negative components of the TGD1st kernel. A small value should be selected for $\delta$, such as $1 \times 10^{-4}$. The TGD1st kernel is characterized by three AKPs: $\theta$ dictates the orientation, while $\gamma_1$ and $\gamma_2$ define the shape.
	
	\textbf{TGD2nd kernel function}
	When the exponential function is employed as the kernel function for second-order directional TGD operator\cite{tao2023tao} , the resultant TGD2nd kernel takes the following form:
	\begin{align}
			&& \mathit{TGD2nd}(x,y;\theta, \gamma) &=  W - sum(W) \times \mathbf{I}_{x=y=0} \\
	where: &&  W  &= e^{-\frac{x^{\prime 2}+ y^{\prime 2}}{\gamma^2}} \times \left|\frac{x^{\prime }}{x^{\prime 2}+ y^{\prime 2}}\right| \\
			&& x^{\prime} &= x \cos\theta	+ y \sin\theta \\ 
			&& y^{\prime} &= -x \sin\theta	+ y \cos\theta
	\end{align}	
	Here, $\mathbf{I}_{x=y=0}=1$ if $x=y=0$ otherwise 0.  TGD2nd kernel has two AKPs: $\theta$ and $\gamma$.
	
	\textbf{LoT kernel function}
	LoT(Laplacian of TGD) is defined in \cite{tao2023tao}, here we use exponential function as its kernel function :
	\begin{align}
			&&\mathit{LoT}(x,y; \sigma ) &=  W - sum(W) \times \mathbf{I}_{x=y=0} \\
			where:	&&  W  &= e^{-\frac{x^2+ y^2}{\sigma^2}}
	\end{align}
	LoT kernel has only one AKP, $\sigma$.
	
	Researchers are encouraged to develop and select superior ACKs when better-suited kernel functions emerge for specific application contexts. Such flexibility is inherent to ACL and constitutes the fundamental source of its complexity, ensuring a robust representation in feature space.
	
	\subsection{Aanlytically model pretrained convolution kernels}
	
	\begin{figure}[!tb]
		\centering
		\includegraphics[width=0.9\linewidth]{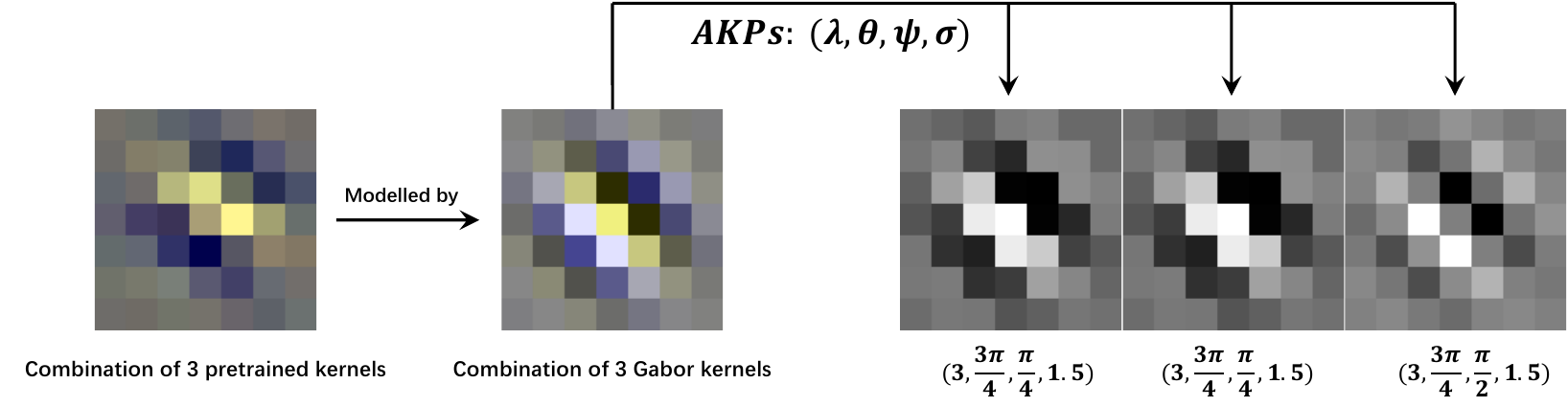}
		\caption{Certain patterns observed in pretrained convolution kernels can be accurately modeled using ACKs. This demonstration illustrates the ability to use significantly fewer AKPs to approximate pretrained kernels. On the left, there is a pattern (which is actually a combination of three kernels) selected from the pretrained weights of ResNet-50. It is precisely modeled by three Gabor kernels, described by a mere 12 AKP values.}
		\label{fig:decom}
	\end{figure}
	Upon establishing kernel functions, these can be harnessed to construct ACKs that precisely mimic those observed in practical settings. \cref{fig:decom} depicts an analytical model of a pretrained kernel (actually three kernels corresponding to the RGB channels ) from the first convolution layer of ResNet-50  found in the third row and first column of \cref{fig:sim}. Using only 12 AKPs, this model faithfully reproduces a pattern from the pretrained kernels, which originally consists of 147 parameters (amounting to three $7\times 7$ kernels). This example highlights the remarkable expressive power of ACKs and their proficiency in facilitating model compression.
	
	\subsection{Make your neural network more explainable}
	To enhance the interpretability of your neural network, or to elucidate the mechanics of specific layers or kernels, simply substitute your convolutional layers with ACLs to transition to an \textbf{Ana}lytic \textbf{N}eural \textbf{N}etwork (AnaNN). The primary considerations in this process are the selection of appropriate kernel functions and the arrangement of ACKs within  ACLs. Alternatively, should you prefer not to employ an AnaNN, you can still utilize the AKPs  honed by the AnaNN to forge ACKs. These can then be amalgamated with plain kernels to assemble the full complement of kernels required for traditional convolutional layers. Subsequently, integrating these kernels into your extant neural network architecture will not only bolster reliability but also improve interpretability, as a significant proportion of the convolution kernels are derived from explicit kernel functions. This methodology is illustrated in \cref{fig:AnaNN}.

	\section{Experiments}
	\label{sec:exp}

	In this section, extensive experiments are conducted to assess the capabilities of the Analytic Convolutional Layer (ACL) and to evaluate the practicality of implementing Analytic Neural Networks (AnaNNs). Throughout our experimentation, a variety of kernel types were employed, including  Gabor (G), LoG (Lg), LoT (Lt), TGD1st (Tf), TGD2nd (Ts), Mean (M), and Plain (P, \ie traditional convolution kernels) kernels. To quantify the parameter efficiency afforded by an ACL, we introduce the compact factor, which calculates the proportion of learnable parameters reduced by utilizing an ACL:
	\begin{equation}
		\begin{aligned}
			\mathit{compact\ factor} = 1 - \frac{\sum p_{i}k_{i} + n \times h \times w}{C_{o}\times C_{i}\times h \times w}\\
		\end{aligned}
	\end{equation}
	Here, $k_{i}$ represents the number of ACKs of the $i$th type, $p_{i}$ denotes the number of corresponding AKPs per ACK, and $n$ signifies the number of Plain kernels in the ACL. The balance between the ACL's compactness and its performance capability rests in your hands. 
	
	\subsection{ACL tests}
	\begin{figure}[!htb]
		\centering
		\subfloat[Initialization of an ACL]{ 
			\includegraphics[width=0.45\linewidth]{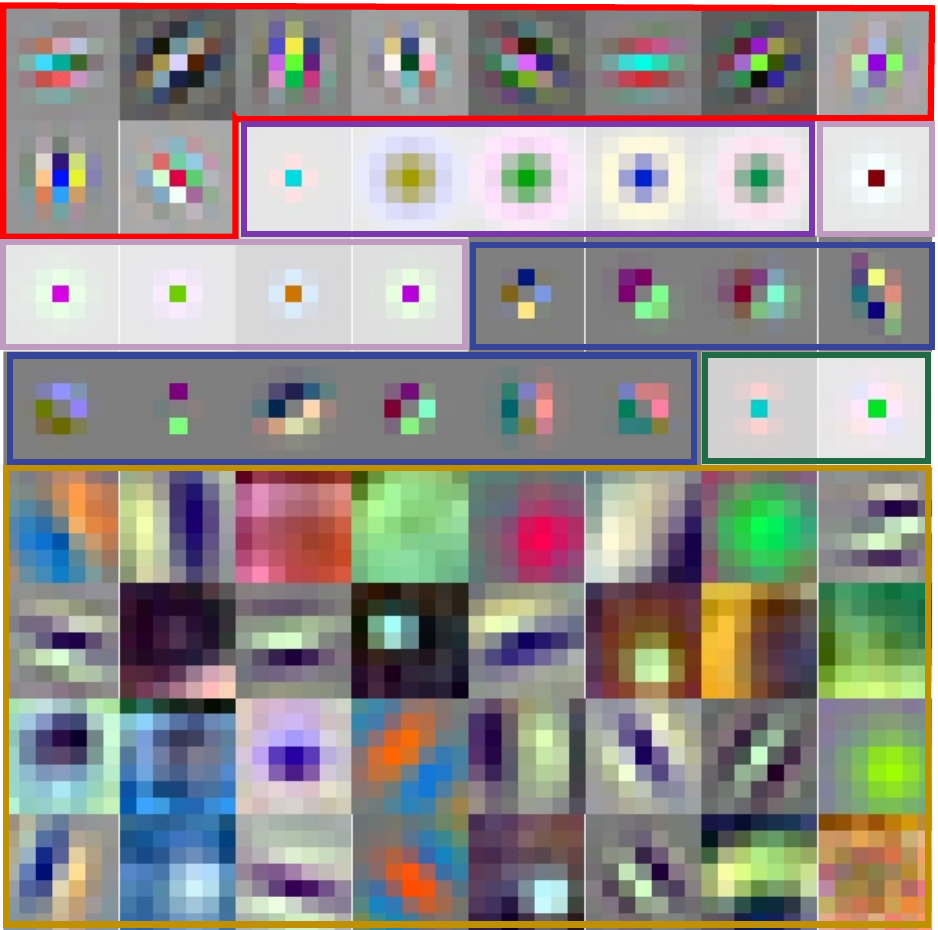}
			\label{fig:init}
		} {\color{white}}
		\hfill
		\subfloat[Top-1 accuracy contributions explanation]{ 
			\includegraphics[width=0.45\linewidth]{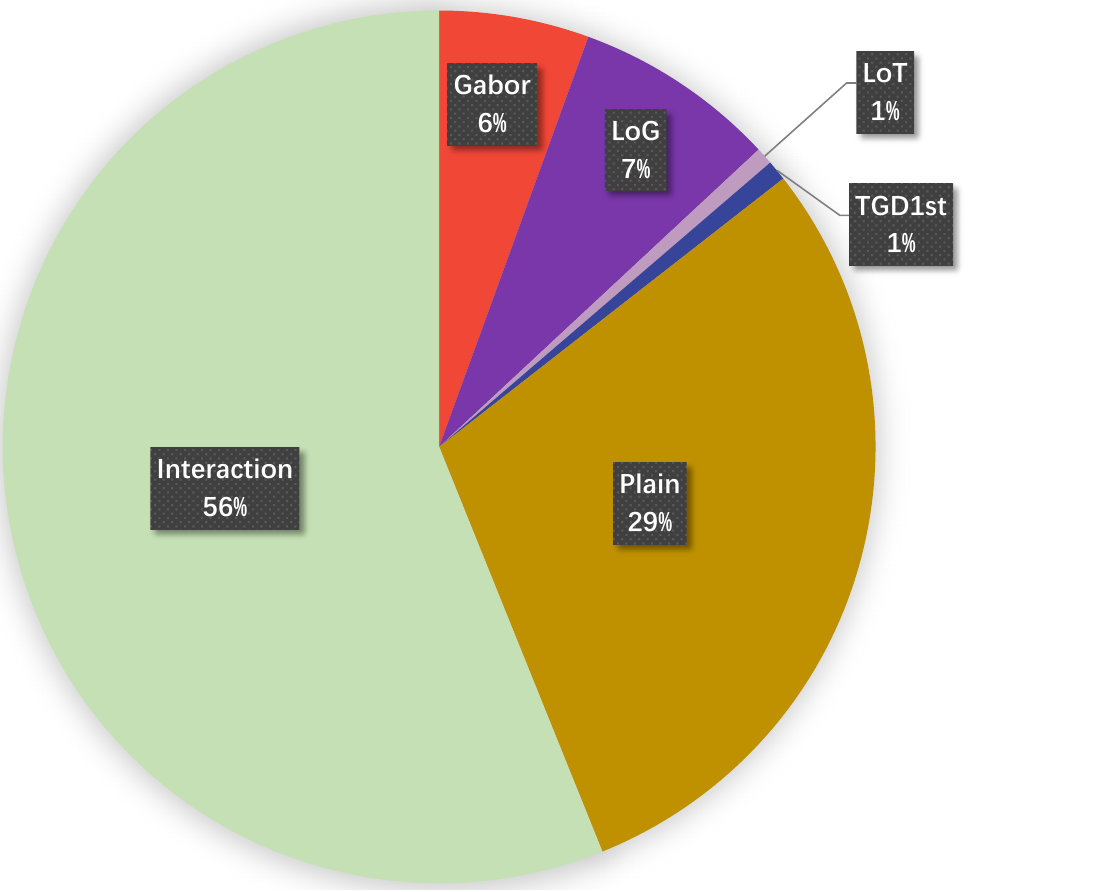}
			\label{fig:explain}
		} {\color{white}}
		\caption{(a): Visualization of an ACL mosaic in its initialization state  with the arrangement $(3\times 64)\mathit{G}_{30}\mathit{Lg}_{15}\mathit{Lt}_{15}\mathit{Tf}_{30}\mathit{Ts}_{6}\mathit{P}_{96}$. The kernels framed by red, purple, pink, blue, green, and orange outlines correspond to Gabor, LoG, LoT, TGD1st, TGD2nd, and Plain kernels, respectively. ACKs are initialized by random AKP values, while Plain kernels are initialized with pretrained kernels. They are reorganized into a $ 64 \times 3 $ shape for display in RGB mode. (b): Top-1 accuracy contributions explanation of the various kernel types in exp3 elucidated by an ablation study.}
		\label{fig:exp}
	\end{figure}
	Building on the empirical observation that distinct patterns arise from lower-layer convolution kernels, our study embarked on exploring how effectively ACKs could replicate these traditional convolution kernels. For this purpose, the Oxford 102 Flower dataset\cite{Nilsback08} was selected, comprising 102 flower categories, with each category containing between 40 and 258 images. The training and validation datasets each consist of 1020 images, whereas the test dataset includes 6149 images. We utilized a pretrained ResNet-34\cite{He2015DeepRL} model as our baseline. All experiments conducted, and represented in \cref{tab:trade}, \cref{tab:order}, and \cref{tab:ablation}, adhered to the same network architecture as the standard ResNet-34, with two key modifications: the first convolutional layer was replaced by an ACL, and the fully-connected layer was substituted with a Fastai-styled \cite{howard2018fastai} header (same for all experiments) to amplify accuracy. The ACL comprised 192 kernels, accounting for an input and output channel count of 3 and 64, respectively, and kernel dimensions of $7\times 7$. Consequently, the ACL contained at most 9408 parameters subject to optimization via the backpropagation (BP) algorithm. Over the course of the experimentation, networks were trained for 300 epochs with the pinnacle top-1 accuracy on the test dataset being documented. The Fastai library was used throughout the training phase to sustain model generalization performance and preclude overfitting. An illustrative depiction of an ACL in its initialization state is provided in \cref{fig:init}.
	
	\begin{table}[htb]
		\centering
		\caption{Trade-off between feature characterization capabilities and compactness}
		\label{tab:trade}
		\begin{tabular}{c|c|c|c}
			\toprule
			Exps &     kernels assignment                     & Top-1 Acc(\%)         &  compact factor\\ 
			\midrule
			exp1 & $\mathit{P}_{192}$                                  & \textbf{91.79}          &   \textbf{0.}     \\
			\midrule
			exp2 & $\mathit{G}_{30}\mathit{Lg}_{15}\mathit{Lt}_{15}\mathit{Tf}_{30}\mathit{Ts}_{6}\mathit{P}_{96}$  &  90.39           &   0.4732 \\
			\midrule
			exp3 & $\mathit{G}_{30}\mathit{Lg}_{15}\mathit{Lt}_{15}\mathit{Tf}_{36}\mathit{P}_{96}$        & \textbf{ 90.68}  &   \textbf{0.4726} \\
			\midrule
			exp4 & $\mathit{G}_{72}\mathit{Lg}_{30}\mathit{Lt}_{15}\mathit{Tf}_{27}\mathit{P}_{48}$       &   87.74          &  0.706  \\
			\midrule
			exp5 & $\mathit{G}_{60}\mathit{Lg}_{24}\mathit{Lt}_{27}\mathit{Tf}_{45}\mathit{Ts}_{36}$      &  86.10          &   0.9471 \\
			\midrule				
			exp6 & $\mathit{G}_{120}\mathit{Lg}_{30}\mathit{Lt}_{15}\mathit{Tf}_{27}$             &  85.43          &   0.9356  \\
			\midrule
			exp7 & $\mathit{M}_{192}$                                   &  73.39          &  1.  \\   
			\bottomrule
		\end{tabular}
		
	\end{table}
	The trade-off between representational capacity and compactness is a fundamental challenge in neural network design. Unlike other models that aim for compactness, the ACL allows for a nuanced balance between these aspects, as demonstrated by the experiments recorded in \cref{tab:trade}. Our initial experiment, exp1, attained an accuracy of $91.79\%$; it's worth noting that an ACL comprised entirely of plain kernels is tantamount to a traditional convolutional layer. However, as the compact factor increases, we observe a decrement in top-1 accuracy, implying a necessary compromise. In exp3, a $90.68\%$ top-1 accuracy was achieved while saving $47.26\%$ of parameters. This outcome is significant, especially when considering that ResNet-152-SAM\cite{chen2021vision} obtained a $91.1\%$ accuracy and ResNet-50-SAM garnered $90\%$\cite{chen2021vision}. With a comparatively minimal number of parameters, we obtained a respectable level of classification accuracy. The fact that the ACL was only applied to the first convolutional layer and still yielded such promising results is quite encouraging.
	
	The relationship between decreased top-1 accuracy and increased model compactness is not strictly linear. A vital contributing factor to this is the ordering of ACKs, assuming their types remain constant. This was evidenced by the findings presented in \cref{tab:order}, which illustrates that the performance of the model is intricately linked to the sequence of ACKs deployed.
	
	\begin{table}[!htb]
		
		\centering
		\caption{The order of ACKs matters}
		\label{tab:order}
		\begin{tabular}{c|c|c|c}
			\toprule
			Exps &     kernels assignment                     & Top-1 Acc(\%)        &  compact factor\\ 
			\midrule
			exp3 & $\mathit{G}_{30}\mathit{Lg}_{15}\mathit{Lt}_{15}\mathit{Tf}_{36}\mathit{P}_{96}$        & 90.68          &   0.4726 \\
			\midrule
			exp7 & $\mathit{Lg}_{15}\mathit{G}_{30}\mathit{Lt}_{15}\mathit{Tf}_{36}\mathit{P}_{96}$        & 90.26         &   0.4726 \\
			\midrule
			exp8 & $\mathit{Lg}_{15}\mathit{G}_{30}\mathit{P}_{96}\mathit{Lt}_{15}\mathit{Tf}_{36}$        & 89.22         &   0.4726 \\
			\bottomrule
		\end{tabular}
	\end{table}
	To investigate the distinct roles that each type of ACK plays in feature representation, we conducted ablation experiments by substituting each ACK type with Mean kernels to determine the extent of impact on the ACL’s capacity. The outcomes of these experiments are delineated in \cref{tab:ablation}. A comparison between exp3 and exp15 reveals a significant top-1 accuracy decrease of $17.29\%$ when all ACKs are replaced by Mean filters. To elucidate the contribution of the five kernel types to top-1 accuracy, individual experiments (exp9 to exp13) replaced each kernel type with Mean kernels to observe the resultant accuracy reduction. These decrements in performance are thought to approximately signify the contributions of the respective kernel types, as visualized in the pie plot in \cref{fig:explain}. Notably, we attribute an accuracy loss of about $9.7\%$ to the synergistic interactions (or combinations) among the ACKs, accounting for $56\%$ of the contributory significance evidenced in exp3. Subsequent experiments (exp16 to exp19) reinforce the conclusion that it is the interplay between different ACKs, rather than the individual kernels themselves, that is primarily responsible for the model’s capability to represent the feature space effectively.
	
	\begin{table}[!htb]
		\centering
		\caption{Ablation study}
		\label{tab:ablation}
		\begin{tabular}{c|c|c|c|c}
			\toprule
			Exps  &     kernels assignment                     & Top-1 Acc(\%) &  compact factor  & Acc loss (\%)\\ 
			\midrule
			exp3  & $\mathit{G}_{30}\mathit{Lg}_{15}\mathit{Lt}_{15}\mathit{Tf}_{36}\mathit{P}_{96}$        &  90.68        &  0.4726          & 0.  \\
			\midrule
			exp9  & $\mathit{M}_{30}\mathit{Lg}_{15}\mathit{Lt}_{15}\mathit{Tf}_{36}\mathit{P}_{96}$        &  89.72        &  0.4853          & 0.96    \\
			\midrule
			exp10 & $\mathit{G}_{30}\mathit{M}_{15}\mathit{Lt}_{15}\mathit{Tf}_{36}\mathit{P}_{96}$         &  89.38        &   0.4743         & 1.3     \\
			\midrule
			exp11 & $\mathit{G}_{30}\mathit{Lg}_{15}\mathit{M}_{15}\mathit{Tf}_{36}\mathit{P}_{96}$         &  90.57        &   0.4347         & 0.11     \\
			\midrule
			exp12 & $\mathit{G}_{30}\mathit{Lg}_{15}\mathit{Lt}_{15}\mathit{M}_{36}\mathit{P}_{96}$         &  90.55        &   0.4841         & 0.13    \\
			\midrule
			exp13 & $\mathit{G}_{30}\mathit{Lg}_{15}\mathit{Lt}_{15}\mathit{Tf}_{36}\mathit{M}_{96}$        &  85.09        &   0.0274         & 5.09   \\
			\midrule
			exp14 & $\mathit{M}_{96}\mathit{P}_{96}$                             &  88.34        &   0.5            & 2.34   \\
			\midrule
			exp15 & $\mathit{M}_{192}$                                  &  73.39        &  1.            & 17.29\\
			\midrule
			exp16 & $\mathit{G}_{192}$                                  &  85.69        &  0.9184         &  \textbf{\textemdash}  \\
			\midrule
			exp17 & $\mathit{Lg}_{192}$                                 &  84.47        &  0.9796           &  \textbf{\textemdash}  \\
			\midrule
			exp18 & $\mathit{Lt}_{192}$                                 &  82.32        &  0.9796            &  \textbf{\textemdash}  \\
			\midrule
			exp19 & $\mathit{Tf}_{192}$                                 &  84.00        &  0.9388          &  \textbf{\textemdash}  \\
			\bottomrule
		\end{tabular}
	\end{table}

	\subsection{AnaNN tests}
	\begin{figure}[!tb]
		\centering
		%		\captionsetup[subfloat]{labelsep=none,format=plain,labelformat=empty}
		\subfloat[AnaNN-LeNet]{ 
			\includegraphics[width=0.35\linewidth]{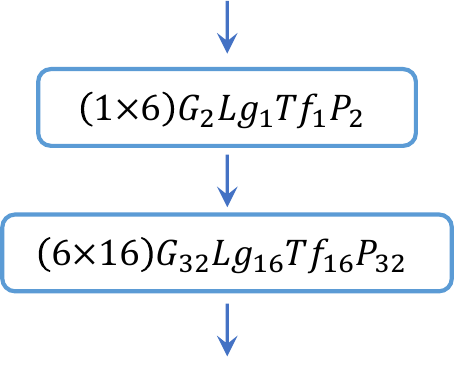}
			\label{fig:anaLenet}
		} {\color{white}}
		\subfloat[Four-layer AnaNN]{ 
			\includegraphics[width=0.45\linewidth]{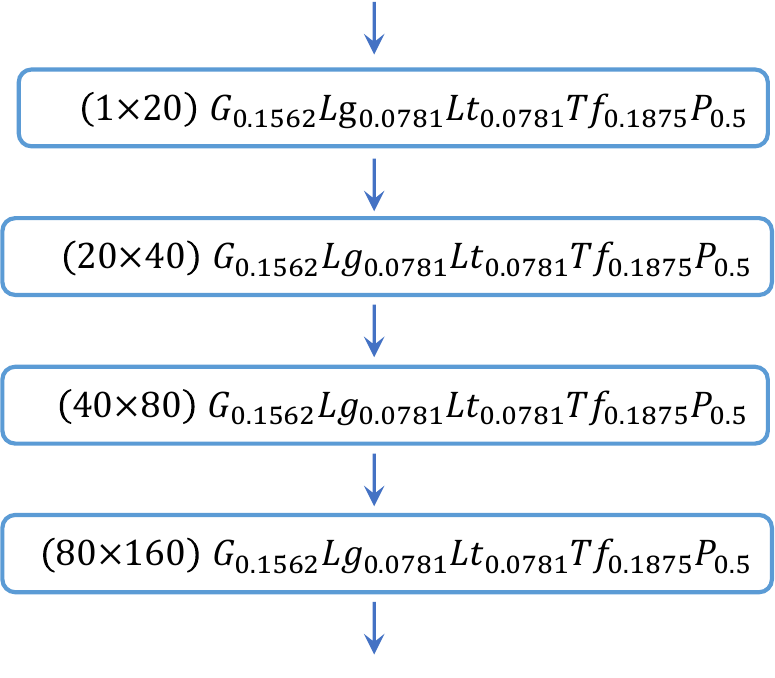}
			\label{fig:MyAnaNN}
		} {\color{white}}
		\caption{a: the structure of AnaNN-LeNet. b: our four-layer AnaNN. Other network modules omitted for simplicity. }
		\label{fig:acconv}
	\end{figure}
	
	We assessed several Convolutional Neural Networks (CNNs) and corresponding Analytic Neural Network (AnaNN) variants using the MNIST dataset \cite{lecun1998mnist}, which contains 60,000 handwritten images in the training set and 10,000 in the test set. Our baseline model, a standard LeNet \cite{lecun1998gradient}, achieved a $96.02\%$ accuracy. After replacing its two convolutional layers with ACLs, we obtained the AnaNN-LeNet variant, which reached an impressive $98.49\%$ top-1 accuracy, surpassing the baseline by $2.47\%$. This model's structure is illustrated in \cref{fig:anaLenet}. Subsequently, we developed a four-layer CNN that realized a $99.58\%$ accuracy on the MNIST dataset. When its four traditional convolutional layers were substituted with ACLs, using a consistent ratio of kernel assignments $\mathit{G}_{0.1562}\mathit{Lg}_{0.0781}\mathit{Lt}_{0.0781}\mathit{Tf}_{0.1875}\mathit{P}_{0.5}$, same as exp3, as depicted in \cref{fig:MyAnaNN}, this enhanced four-layer AnaNN model slightly outperformed its baseline counterpart with a $99.59\%$ top-1 accuracy. The outcomes of these experiments underscore the potential of AnaNNs for future applications.

	\section{Conclusion}
	\label{sec:conclusion}
	In this study, through the analytical modeling of convolution kernels, we have developed a new model-driven convolutional layer, which is a mosaic of ACKs and some traditional convolution kernels. ACKs are constructed on a solid mathematical foundation and designed to capture distinct facets in the feature space. The parameters of ACKs are learned and updated via the backpropagation algorithm. This approach endows ACLs with a capacity to represent features effectively. Furthermore, the utilization of AKPs reduces the number of parameters that need to be learned, and the compactness of the model can be adjusted according to the requirements.
	
	%	\subsubsection{Complexity}
	\textbf{Complexity}
	The computational complexity of ACL is comparable to that of traditional convolutional layers and can be further reduced through optimization from the ground up. For instance, integrating commonly used kernel functions, such as Gabor kernels, directly into CUDA operators can lead to efficiency gains. The intrinsic complexity of ACL primarily hinges on two factors: the design and selection of ACKs tailored to specific downstream applications, and the strategic arrangement of these ACKs to optimize performance when the compact factor is fixed. Per the well-known no free lunch theorem \cite{wolpert1996lack, 585893}, this inherent complexity is crucial for ensuring the capability of ACL. Nevertheless, empirical guidance can be gleaned from a moderate amount of experimentation, for example, the kernel assignment ratio utilized in exp3 has proven to be a viable choice.
	
	\textbf{Neural Network Interpretation}
	Our experimental evidence indicates the existence of complex higher-order mechanisms within convolution layers. Thanks to the model-driven design of ACLs, we possess the capability to evaluate the functionality of distinct network elements with precision and in quantitative terms. The ablation studies conducted present a systematic approach for discerning the underlying contributors to model expressivity afforded by ACLs. These results lay the groundwork for a solid analytical framework that probes the intricate dynamics of neural networks, thereby opening avenues for groundbreaking exploration in interpreting neural network.

	\bibliographystyle{style/splncs04}
	\bibliography{bib/bmain.bib}

%\appendix

\chapter*{Supplementary materials}

The space limit of this paper precludes a detailed description of all relevant experiments within the main text. Our study primarily examines the architecture of Analytic Convolutional Layers (ACLs), while the supplementary material provides evidence of their versatility across various neural networks and datasets.We conducted tests on a suite of convolutional neural networks (CNNs), including DenseNet121\cite{huang2017densely}, ResNet18\cite{He2015DeepRL}, GoogleNet\cite{szegedy2015going}, EfficientNet-B0\cite{tan2019efficientnet}, EfficientNetV2\_L\cite{tan2021efficientnetv2}, MobileNetV2\cite{howard2017mobilenets}, RegNetX\_400MF\cite{radosavovic2020designing}, VGG16\cite{simonyan2014very}, and ShuffleNet\_V2\_X1\_0\cite{ma2018shufflenet}, along with their respective ACL counterparts. To underscore the breadth of our work and its practical significance, we will publicly release the source code for our experiments, facilitating replication and underscoring our commitment to methodological transparency. The supplementary materials consist of a detailed PDF covering experiments not included in the main text, the ACL source code, and additional resources necessary for reproducing the experiments.

\section{Supporting Experiments} 
This paper focuses on exploring the properties of Analytic Convolutional Layers (ACLs). Supplementary experiments detailed here support our assertion that the findings are broadly applicable and that ACLs consistently demonstrate robust performance on a wide range of natural image datasets and different neural network architectures.

In subsequent sections, we adopt the convention of prefixing a model with \emph{ACL-} when its first convolutional layer has been replaced by an ACL. Models with multiple convolutional layers substituted by ACLs are denoted with the prefix \emph{AnaNN-}.

\subsection{ACL Tests}
\subsubsection{Food-101 Dataset}

To further affirm the efficacy of Analytic Convolutional Layers (ACL), we employed ACL-DenseNet121 for a classification task on the Food-101 dataset \cite{bossard2014food}. This dataset consists of 101 food categories, each providing 750 training images and 250 test images, totaling 101,000 images overall. The baseline ACL-DenseNet121 configuration, denoted as $(3\times 64)P_{192}$, mirrors the architecture of the standard DenseNet121 , with the exception of the fully-connected layer, which is replaced by a custom header. This baseline model achieved an accuracy of $84.93\%$ within 100 epochs. An enhanced ACL-DenseNet121 version with the ACL represented as $(3\times 64)\mathit{G}_{30}\mathit{Lg}_{15}\mathit{Lt}_{15}\mathit{Tf}_{36}\mathit{P}_{96}$, equivalent to \emph{exp3} mentioned in the paper, and featuring a compactness factor of $47.26\%$, reached an accuracy of $85.01\%$, sligfhtly  outperforms the standard DenseNet121 . In comparison, the \emph{vit-base-food101-demo-v5} model \footnote{\href{https://huggingface.co/eslamxm/vit-base-food101}{https://huggingface.co/eslamxm/vit-base-food101}} achieved an accuracy of $85.39\%$ on the Food-101 dataset, underscoring the competitive performance of our ACL-DenseNet121 model.

\subsubsection{CIFAR-10 Dataset} We extended our evaluation of ACL-enhanced networks to the CIFAR-10 dataset \cite{krizhevsky2009learning}, which is composed of 60,000 32x32 color images distributed across 10 classes, with each class containing 6,000 images. The dataset is split into 50,000 training images and 10,000 test images. The baselines in this context refer to the original versions of various networks, such as ResNet18, GoogleNet, VGG16, EfficientNet-B0, EfficientNetV2\_L, MobileNetV2, and RegNetX\_400MF. The primary objective was to demonstrate that ACL-based networks maintain robust performance while significantly reducing the number of learned parameters. We did not employ complex techniques to align our baseline experiments with state-of-the-art networks; rather, we ensured identical conditions for the baseline networks and their corresponding ACL-modified counterparts.

For all ACL-related tests, the layers maintained the same kernels ratio, $\mathit{G}_{0.1562}\mathit{Lg}_{0.0781}\mathit{Lt}_{0.0781}\mathit{Tf}_{0.1875}\mathit{P}_{0.5}$, aligning with exp3 from the paper. The results of these experiments are presented in \cref{tab:aclTest}. It is evident that the ACL-networks preserve a high level of accuracy despite a 47.16\% reduction in the number of learnable parameters; in certain instances, as highlighted by the bolded figures in \cref{tab:aclTest}, they even surpass the performance of the original networks.

\begin{table}[!htb]
	\centering
	\caption{Different ACL- networks tested on CIFAR10 dataset}
	\label{tab:aclTest}     
	\begin{tabular}{c|c|c|c}
		\toprule
		Network    & Baseline Acc(\%) &  Top-1 Acc  & compact factor (\%) \\ 
		\midrule
		ACL-ResNet18 &   92.24       &  91.92       & 47.26  \\
		\midrule
		ACL-GoogleNet  &   91.31    & 90.28  &  47.26 \\
		\midrule
		
		ACL-EfficientNet-B0 &  84.96 & \textbf{86.34} & 47.26 \\
		\midrule
		ACL-MobileNetV2    &  86.42 & 85.32 & 47.26 \\
		\midrule
		ACL-RegNetX\_400MF  &  80.56   &   80.48     & 47.26 \\
		\midrule
		ACL-VGG16          &   91.60     &   \textbf{92.44}   &     47.26       \\
		\midrule
		ACL-ShuffleNet\_V2\_X1\_0  & 82.07 & 80.15 & 47.26 \\
		\midrule
		ACL-EfficientNetV2\_L & 93.15  & 92.81 &  47.26 \\
		\bottomrule
	\end{tabular}
\end{table}

\subsection{AnaNN Tests}
\begin{figure}[!tb]
	\centering
	%		\captionsetup[subfloat]{labelsep=none,format=plain,labelformat=empty}
	
	\includegraphics[width=0.8\linewidth]{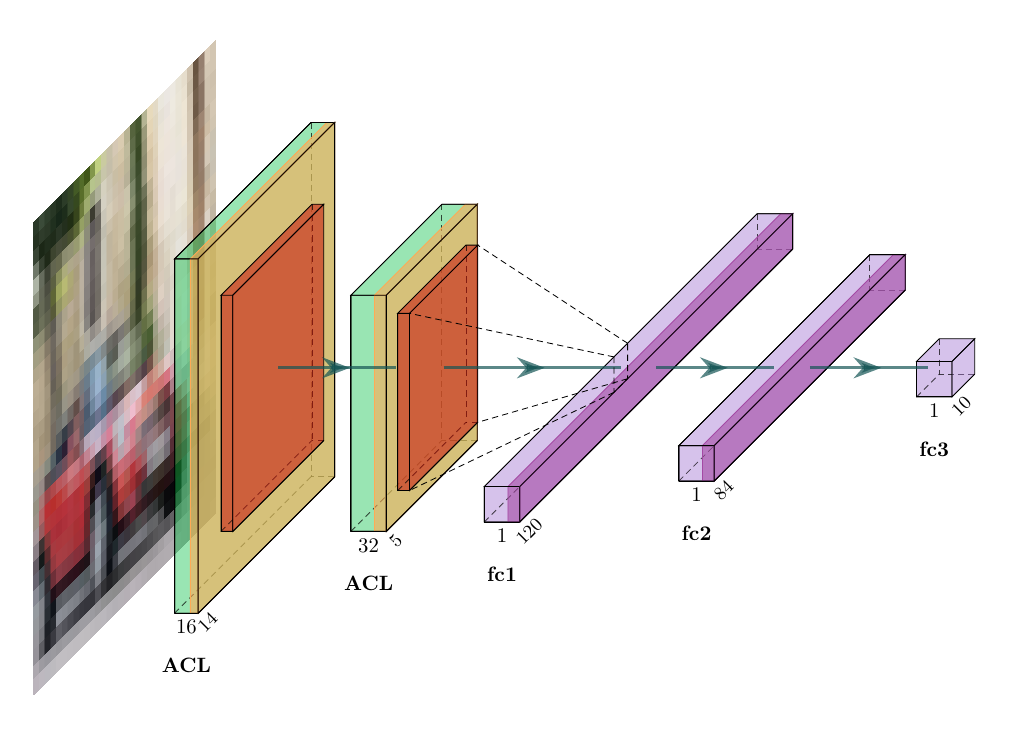}
	\caption{The structure of AnaNN-LeNet, whose two ACLs have the same ratio form with exp3, $\mathit{G}_{0.1562}\mathit{Lg}_{0.0781}\mathit{Lt}_{0.0781}\mathit{Tf}_{0.1875}\mathit{P}_{0.5}$.}
	\label{fig:AnaNN-LeNet}
\end{figure}

A neural network is termed an AnaNN (analytic neural network) when it incorporates Analytic Convolutional Layers (ACLs) in more than one of its convolutional layers. In this study, the AnaNN-LeNet model was employed for classification tasks on the CIFAR-10 dataset, and its architecture is illustrated in \cref{fig:AnaNN-LeNet}. The conventional LeNet baseline attained an accuracy of $73.48\%$. Meanwhile, our AnaNN-LeNet model achieved an accuracy of $71.18\%$. Despite a slight decrease in accuracy, the AnaNN-LeNet model resulted in a $47.26\%$ reduction of parameters within its convolutional layers.

\section{Source Code and Replication of Experiments}
	 Our implementation of Analytic Convolutional Layer (ACL) adheres strictly to PyTorch\cite{NEURIPS2019_9015} conventions, allowing for seamless substitution of standard convolutional layers with ACLs in any neural network architecture. This flexible compatibility with a range of architectures has been demonstrated through our comprehensive experiments. Moreover, the advent of ACL heralds a new era for crafting analytic and transparent networks, significantly enhancing the interpretability of neural network models. Further investigation into designing superior models for the kernel functions and developing more capable ACLs is necessary and requires additional statistical analysis and experimental validation. To aid researchers in replicating our studies and contributing to the development of more transparent and interpretable neural networks, we are providing our ACL source code and experiment replication materials as supplementary attachments.
	 
% \bibliographystyle{../style/splncs04}
% \bibliography{supplementary.bib}

\end{document}